\documentclass[cameraready]{Interspeech}

\title{Listening or Reading? Evaluating Speech Awareness \\ in Chain-of-Thought Speech-to-Text Translation}

\author[affiliation={1}, orcid=0009-0009-1503-7045]{Jacobo}{Romero-Díaz}
\author[affiliation={1,2}, orcid=0000-0001-7466-3606]{Gerard I.}{Gállego}
\author[affiliation={1}, orcid=0009-0001-4084-1598]{Oriol}{Pareras}
\author[affiliation={1,2}, orcid=0000-0002-1389-3595]{Federico}{Costa}
\author[affiliation={1,2}, orcid=0000-0002-1730-8154]{\\Javier}{Hernando}
\author[affiliation={1,3}, orcid=0000-0001-5414-4710]{Cristina}{España-Bonet}

\address{
    $^1$ Barcelona Supercomputing Center, AI Institute, Spain \\
    $^2$ Universitat Politècnica de Catalunya, Spain \\
    $^3$ DFKI GmbH, Saarland Informatics Campus, Germany
}

\email{gerard.ion.gallego@upc.edu, \{oriol.pareras;federico.costa\}@bsc.es}


\usepackage{comment}

\usepackage{amsmath}
\usepackage{graphicx}
\usepackage{hyperref}
\usepackage{comment}
\usepackage{siunitx}
\usepackage{multirow}
\usepackage{float}
\usepackage[most]{tcolorbox}
\usepackage{lipsum}
\usepackage{nth}
\usepackage{booktabs}
\usepackage{textgreek}

\usepackage{xspace}

\usepackage{tabularray}
\UseTblrLibrary{booktabs}
\definecolor{HeatColor}{RGB}{186,224,186}
\usepackage{threeparttable}

\newcommand{\vsix}{\textsc{Base}\xspace}
\newcommand{\vseven}{\textsc{Dual}\xspace}
\newcommand{\veightninetenbase}{\textsc{Noisy}}
\newcommand{\veightnineten}{\veightninetenbase\xspace}
\newcommand{\veight}{\veightninetenbase\xspace}

\newcommand{\xcomet}{x\textsc{comet}\xspace}
\newcommand{\xcometxl}{x\textsc{comet-xl}\xspace}

\newcommand{\direct}{\textsc{Direct}\xspace}
\newcommand{\coft}{\textsc{CoT}\xspace}
\newcommand{\cascade}{\textsc{Cascade}\xspace}

\begin{document}

\maketitle

\begin{abstract}
    Speech-to-Text Translation (S2TT) systems that cascade Automatic Speech Recognition (ASR) and Text-to-Text Translation (T2TT) are fundamentally limited by error propagation and the loss of prosodic cues. Chain-of-Thought (CoT) prompting was introduced to address these limitations by providing joint access to both speech and transcription during generation. In this study, we systematically evaluate CoT's actual speech awareness through input attribution analysis, assessment of robustness against corrupted transcripts, and prosody-awareness evaluation. Our results indicate that CoT largely replicates cascaded behavior, relying heavily on textual transcripts. By applying simple training interventions (e.g., injecting noisy transcripts in the CoT), we demonstrate enhanced model robustness and increased acoustic reliance. Our findings challenge the presumed benefits of CoT S2TT and emphasize the necessity for techniques that help integrate acoustic information into the translation process.
\end{abstract}

\section{Introduction}

Speech-to-Text Translation (S2TT) refers to converting spoken utterances in one language into written text in another.
Historically, S2TT systems were built by cascading an Automatic Speech Recognition (ASR) module, which first transcribes the speech, and a Text-to-Text Translation (T2TT) module which translates it into the target language~\cite{cascade_st}.
Such \emph{cascade} systems are simple but have two main limitations. First, they can suffer from error propagation, as the T2TT stage cannot correct transcription errors. Second, they cannot preserve information in speech that can be relevant for translation, such as prosody, including emphasis, or breaks that convey a certain meaning~\cite{sperber-paulik-2020-speech}.

To address these limitations, alternative approaches have been proposed. In \emph{Direct} S2TT models, speech is directly translated without intermediate text representations~\cite{berard2016}. However, these systems tend to underperform cascade models, mostly due to the scarcity of large-scale parallel S2TT data.
More recently, research has shifted toward Speech Large Language Models (SLLMs), introducing a new paradigm of \emph{Chain-of-Thought} (CoT) or \textit{multi-turn} S2TT~\cite{huang2023speech,hu2025chain}. Similar to cascade systems, these models first generate the transcription and then produce the translation. The key difference is that they retain access to both the speech input and the transcription during translation. This design is expected to overcome the limitations of cascade models by improving robustness to error propagation and enabling the use of acoustic cues~\cite{huang2023speech}. However, this assumption has not Okay  been thoroughly validated, and to the best of our knowledge, no prior work has systematically examined whether CoT provides such benefits over cascade S2TT.

In this work, we investigate the hypothesized benefits of CoT models across three complementary perspectives. First, we analyze internal mechanisms of a model with an interpretability method, quantifying the amount of information the model uses from the source speech utterance during translation (\S\ref{sec:value_zeroing}). Second, we evaluate robustness to error propagation by simulating noisy ASR outputs. Measuring translation quality under such conditions provides indirect evidence of whether the model exploits speech information to compensate for degraded transcripts (\S\ref{sec:errorprop}). Third, we assess prosody awareness of the model using the \textsc{ContraProST}~\cite{contraprost} benchmark (\S\ref{subsec:prosody}), which directly tests whether the model leverages acoustic cues beyond the transcript to disambiguate meaning when prosody affects translation. We complement these analyses with simple training strategies to encourage greater reliance on speech information during translation on CoT setups (\S\ref{subsec:models}).

\section{Methodology}
\label{sec:methods}

This section describes how we analyze speech awareness in CoT S2TT. We assess their direct contribution to translation through attribution analyses, their indirect role via robustness and prosody evaluations, and compare different model variants to understand how training strategies shape CoT behavior.

\subsection{Models}
\label{subsec:models}

\noindent\textbf{Architecture}\;~ We carry out our study on models consisting of a Speech LLM architecture. Specifically, we extend an LLM-based translation model to accept spoken inputs and perform S2TT. Following recent approaches~\cite{speechgpt,audiopalm}, speech is discretized and incorporated into the model as additional tokens in its vocabulary. Each utterance is first encoded with a self-supervised (SSL) speech model to obtain continuous representations, which are then quantized with k-means clustering into \emph{discrete speech units} (DSUs). The embedding layer of the base LLM is expanded to accommodate these tokens. After this discretization, training proceeds as in a text-only LLM.

\noindent\textbf{Inference Strategies}\;~ We compare the behavior of models under the \textbf{\coft} and the \textbf{\cascade} strategies. Models are trained on a mixture of ASR, T2TT, and S2TT data (\S\ref{sec:data}), which enables them to operate in both modes. In the \coft setup, the model generates a transcription followed by a translation while retaining the original speech input in the context. In the \cascade setup, we adopt a \emph{self-cascade} approach: the model first generates a transcription from the speech input and then, conditioned solely on this transcription, produces the translation.

\noindent\textbf{Training Configurations}\;~ We first consider a \textbf{\vsix} model trained with S2TT data exclusively in the CoT format. This serves as the baseline for comparing the two inference strategies. We hypothesize there will be minimal differences between them, implying that \coft provides no meaningful gains over \cascade. To address this limitation, we introduce two interventions. The first is \textbf{\vseven}, a model trained on a mixture of CoT (25\%) and Direct (75\%) formats. Including samples where the model relies only on the DSUs (Direct) encourages greater awareness of the spoken signal during CoT generation. The second is \textbf{\veightnineten}, a novel training strategy in which, for some CoT samples, we replace the transcript with a noisy variant. This simulates error propagation during training and discourages over-reliance on transcriptions. We follow the same corruption strategy used in the analysis of robustness to transcription errors (\S\ref{sec:errorprop}), and we modify 25\% of the training samples (selected by manual tuning). For these samples, we exclude the transcription from the loss to prevent ASR degradation.

\subsection{Attribution scores}
\label{sec:value_zeroing}

We study how the model combines information from different parts of the input using \emph{Value Zeroing}~\cite{value_zeroing}. This method estimates the relative contribution of input tokens to the generated output and is more reliable than directly inspecting attention weights. By applying Value Zeroing, we can observe which modality the model relies on when producing translations.

We use the \texttt{inseq} library~\cite{inseq} to extract token-level contribution scores, represented as token-token interaction matrices. To analyze modality use, we aggregate these scores over spans of tokens. We focus on three input regions: the speech input (DSUs), the transcription, and the previously translated tokens. For each region, we compute its contribution by summing the scores of all tokens in the region and averaging across all output tokens. This yields a normalized distribution over input regions, where contributions sum to 1 (with a small portion attributed to special tokens). We then analyze these aggregated scores across layers of the model to trace how reliance on each modality evolves along the model depth (see Fig.~\ref{fig:input_att}).

\subsection{Robustness to transcription errors}
\label{sec:errorprop}

We analyze robustness to transcription errors as an indirect way of testing whether CoT models exploit speech cues. If the model relies only on the transcript, its performance should degrade when the transcript is corrupted. Conversely, meaningful use of the audio input would mitigate this effect.

Following the suggestion of \cite{hu2025chain}, who propose exploring how CoT behaves when fed with lower-quality ASR transcripts, our controlled corruption procedure provides a novel and reproducible way to simulate such conditions and evaluate robustness systematically. For each transcription in a test set, we replace a contiguous fragment of the transcript with an unrelated one (see Fig.~\ref{fig:prompt}). The replacement preserves the length and grammaticality of the sentence but alters its content entirely, ensuring that the corrupted transcript remains fluent while diverging semantically. We generated the corrupted fragment with Gemini-2.0-Flash~\cite{gemini}. The start position is sampled uniformly, and the fragment length is determined by the target corruption ratio (e.g., 2.5–30\% of the words).

\begin{figure}[t]
    \centering
    \begin{tcolorbox}[
        colback=gray!8, 
        colframe=gray!127, 
        boxrule=0.8pt, 
        title={Transcription corruption example}, 
        coltitle=white, 
        colbacktitle=gray!127, 
    ]

    {\normalsize \textbf{Original sample}}

    \vspace{3pt}

    The \textbf{crust is} about 70 km thick on the near side and 100 km thick on the far side. 

    \vspace{7pt}

    {\normalsize \textbf{Corrupted version}}
    \vspace{3pt}
    
    The \textbf{sun shines} about 70 km thick on the near side and 100 km thick on the far side.
        
    \end{tcolorbox}
    \vspace{-1em}
    
    \caption{Example of corrupted transcription generated with Gemini-2.0-Flash (corruption ratio: 15\%).}
    \label{fig:prompt}
    \vspace{-16pt}
\end{figure}

The corrupted transcripts are then paired with the original speech and inserted into a CoT prompt. Results are evaluated with \xcomet~\cite{guerreiro-etal-2024-xcomet}, a well-known translation metric that predicts human judgments from multilingual encoder representations. By measuring the translation quality drop as corrupted percentages increase, we can quantify the robustness of the model to error propagation.

\subsection{Prosody awareness}
\label{subsec:prosody}

We aim to analyze whether CoT models make use of speech inputs to perform prosody-aware S2TT or instead behave like cascade models, which are effectively blind to prosody. To this end, we evaluate prosody awareness using \textsc{ContraProST}~\cite{contraprost}, a benchmark providing paired utterances differing only in prosodic emphasis. These differences alter meaning and therefore require distinct translations, so poor performance indicates an inability to leverage speech information. For each pair, the model translates both utterances, and each output is scored with \xcomet~\cite{guerreiro-etal-2024-xcomet} against both references, yielding a $2{\times}2$ matrix. From this matrix,~\cite{contraprost} calculates two contrastive quality scores: \emph{Global} and \emph{Directional}. We focus on the \emph{Global} metric, as it provides a stricter evaluation by measuring the percentage of double-contrastive pairs where both translations are preferred over their contrastive references. Additionally, we observed that the original formulation does not account for ties in the \xcomet score, which occur frequently in our experiments. To address this, we assign a score of $0.5$ to each tied comparison. Finally, we average the scores across the multiple prosodic phenomena covered by the benchmark to obtain our final results.

\section{Experimental Setup}
\label{sec:exp_setup}

\subsection{Model details}

Our models are based on \textsc{Salamandra\textbf{TA}}~\cite{salamandrata}, a translation LLM derived from \textsc{Salamandra}~\cite{salamandra} that supports 35 European languages. \textsc{Salamandra} is a Llama-based model, following a decoder-only Transformer architecture and sharing most architectural and training design choices, diverging primarily in its tokenizer design, multilingual data curation strategy, and its focus on European languages. We use the instructed version with 7B parameters.\footnote{\url{https://hf.co/BSC-LT/salamandraTA-7b-instruct}} The SSL speech model we employ to encode spoken inputs is mHuBERT, from TWIST~\cite{twist}.\footnote{\url{https://hf.co/slprl/mhubert-base-25hz}} This version downsamples the temporal resolution to just 25 Hz. DSUs are obtained by clustering the 11th-layer representations into 500 centroids using $k$-means. The speech encoder is frozen during training, and only the LLM parameters are updated.

\begin{figure*}[ht]
    \centering
    \scriptsize
    \vspace{-12 pt}
    \includegraphics[width=\textwidth]{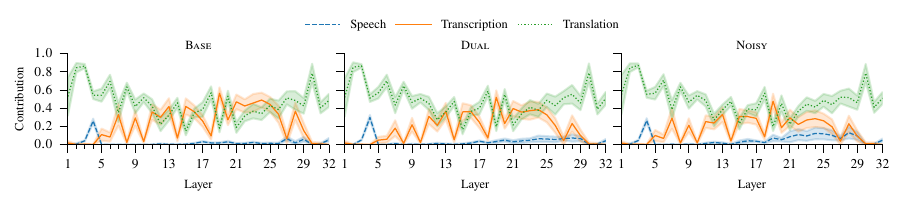}
    \vspace{-3em}
    \caption{Layer-wise attribution scores obtained with \textit{Value Zeroing}, aggregated over \textit{Speech}, \textit{Transcription}, and \textit{Translation} tokens. Each subfigure shows one model variant (\textit{B{\scriptsize ASE}}, \textit{D{\scriptsize UAL}}, \textit{N{\scriptsize OISY}}). Contributions from special tokens are omitted for clarity. Shaded areas indicate the variability across language pairs (mean~$\pm$~standard deviation).}
    \vspace{-6pt}
    \label{fig:input_att}    
\end{figure*}

\begin{figure*}[ht]
    \centering
    \scriptsize
    \vspace{-6pt}
    \includegraphics[width=\textwidth]{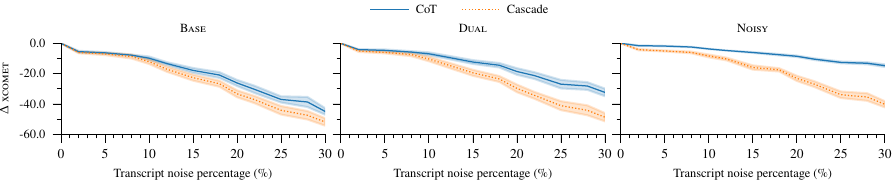}
    \vspace{-1.75em}
    \caption{Robustness to error propagation under controlled transcript corruption. Performance drop ($\mathit{\Delta}$\textit{{\fontsize{5.8pt}{7pt}\selectfont X}{\scriptsize COMET}}) is measured when noisy transcripts are injected into the CoT prompt, relative to ground-truth transcripts (0\% noise percentage). Each subfigure corresponds to one model variant (\textit{B{\scriptsize ASE}}, \textit{D{\scriptsize UAL}}, \textit{N{\scriptsize OISY}}), with curves comparing \textit{C{\scriptsize O}T} and \textit{C{\scriptsize ASCADE}} inference. Results are averaged across languages, and shaded areas indicate the variability across language pairs (mean~$\pm$~standard deviation).}

    \label{fig:errprop}
    \vspace{-9pt}
    
\end{figure*}

\subsection{Data}
\label{sec:data}

\noindent\textbf{Training Mixture}\;~As we extend a T2TT model to perform S2TT, the main focus of the training is to add the speech understanding capability. Because S2TT data are scarce (orders of magnitude below ASR), we train on a mix of ASR and S2TT and keep some T2TT data to avoid catastrophic forgetting. Early experiments revealed that training with external T2TT datasets introduces confounding factors to the study. Therefore, to ensure a fair comparison between \coft and \cascade approaches and avoid potential data quality mismatches, we deliberately reuse the same S2TT data, only using their text transcriptions and translations for the T2TT task. We restrict the T2TT split to $50$k sentence pairs per language, to ensure ASR and S2TT constitute the majority of the data mixture. We employ simple random shuffling of all samples across tasks without any specific scheduling strategy.

\noindent\textbf{Languages}\;~ We train our models to translate from English into six European languages supported by the base translation LLM and with sufficient S2TT data: Catalan (\textsc{ca}), German (\textsc{de}), Spanish (\textsc{es}), French (\textsc{fr}), Italian (\textsc{it}), and Portuguese (\textsc{pt}). To ensure a controlled setup, we restrict source speech to English, thereby minimizing confounding factors such as data scarcity and cross-language imbalance. This also follows the English-to-X setting of \textsc{ContraProST} (\S\ref{subsec:prosody}).

\noindent\textbf{Datasets}\;~We combine several datasets for each task. For ASR, we use the English splits of Common Voice 21.0~\cite{commonvoice} (${\sim}1.6\text{k}$ hours), VoxPopuli~\cite{voxpopuli} (${\sim}500$ hours), and an $8\text{k}$-hour subset of MLS~\cite{pratap20_interspeech}, totaling more than $10\text{k}$ hours. For S2TT, we include CoVoST 2~\cite{covost} (Catalan and German) and Europarl-ST~\cite{europarlst} (all languages except Catalan), contributing an additional ${\sim}500$ hours of speech paired with translations in multiple languages. All evaluations and analyses are conducted on the test subsets of FLEURS~\cite{fleurs} and \textsc{ContraProST}.

\noindent\textbf{Prompt Format}\;~We follow the strategy of the base translation model and format training samples as instructions using OpenAI's \texttt{chatml} template, which casts them into user–assistant interactions. Spoken inputs (DSUs) are represented as strings by mapping them to private-use Unicode characters (PUA), following~\cite{spire}. For ASR and T2TT tasks, the user turn contains the input (DSUs or text) followed by a short instruction (e.g., ``Transcribe in English'' or ``Translate from English into Catalan''), and the assistant turn contains the expected output. For S2TT, we mainly use the CoT strategy, formatting the prompt as multiple turns: the model is first asked to transcribe and then to translate. Unlike a cascade approach, the transcription remains in context during the translation step. In the Direct setup (\S\ref{subsec:models}), S2TT samples are formatted as a single turn that requests the translation directly from the speech input. We train only the assistant outputs, including both turns in the CoT setup.

\vspace{2pt}

\subsection{Training details}

\noindent\textbf{DSU Embeddings Adaptation}\;~To accommodate the new DSU tokens, the base LLM expands its embedding layer (\S\ref{subsec:models}). This layer then combines pretrained text embeddings from the base model with newly initialized embeddings for the DSUs. Prior work suggests adapting these new embeddings to the LLM's representation space before full training~\cite{speechgpt}, which accelerates convergence and improves performance~\cite{gallego25_interspeech}. For this adaptation, we train the model with a next-token prediction objective on discretized speech-only data, keeping the base LLM frozen except for the embedding layer and the output projection.

\noindent\textbf{Hyperparameters}\;~
We train our model with the AdamW optimizer~\cite{loshchilov_decoupled_2019} for $16.7k$ steps with a maximum learning rate (LR) of $1 \cdot 10^{-5}$. The LR follows a cosine schedule, starting with a warmup for the $10\%$ of the training and decaying until $1 \cdot 10^{-6}$. The effective batch size is 256 (16 GPUs $\times$ 16 samples per device). We set a maximum length of $2048$ tokens, and we clip the gradients to a maximum norm of $1.0$. For generation, we use beam-search multinomial sampling, setting 5 beams, a temperature of $0.2$, a top-$p$ of $0.95$, and a top-$k$ of $50$.

\noindent\textbf{Implementation}\;~We use \texttt{transformers}~\cite{wolf_transformers_2020} and Deepspeed for training. We reduce memory usage through mixed precision, gradient checkpointing and Liger Kernel~\cite{hsu_liger_2025}. All trainings are run on NVIDIA H100 GPUs.

\stepcounter{footnote}
\newcounter{tablefnB}
\setcounter{tablefnB}{\value{footnote}}

\begin{table}[ht!]
\centering
\scriptsize
\caption[]{Results on the \textit{C{\scriptsize ONTRA}P{\scriptsize RO}ST} benchmark (\textuparrow), measuring contrastive translation quality (Global score), for three model variants and two generation strategies. Best results are indicated by a more intense {\setlength{\fboxsep}{1pt}\colorbox{HeatColor}{background}}, those with statistically significant\protect\hyperlink{linkB}{\footnotemark[\thetablefnB]} improvements over \textit{B{\scriptsize ASE}} are \underline{underlined}, and \textbf{bold} indicates cases where \textit{C{\scriptsize O}T} significantly outperforms \textit{C{\scriptsize ASCADE}} in the same model variant.}
\begin{tblr}{
  width = 0.35\textwidth,
  colspec = {l *{3}{r}},
  column{1} = {rightsep=16pt},
  column{2-4} = {leftsep=6pt, rightsep=6pt},
  row{1} = {halign=c, font=\scshape, abovesep=2pt, belowsep=1pt},
  row{2} = {abovesep=3pt, belowsep=1pt},
  row{3} = {abovesep=1pt, belowsep=2pt},
  row{4} = {abovesep=3pt, belowsep=1pt},
  row{5} = {abovesep=1pt, belowsep=2pt},
  row{6} = {abovesep=3pt, belowsep=1pt},
  row{7} = {abovesep=1pt, belowsep=2pt},
  cell{2}{2}={bg=HeatColor!0}, cell{2}{3}={bg=HeatColor!9}, cell{2}{4}={bg=HeatColor!0},
  cell{3}{2}={bg=HeatColor!56}, cell{3}{3}={bg=HeatColor!63}, cell{3}{4}={bg=HeatColor!54},
  cell{4}{2}={bg=HeatColor!28}, cell{4}{3}={bg=HeatColor!0}, cell{4}{4}={bg=HeatColor!4},
  cell{5}{2}={bg=HeatColor!89}, cell{5}{3}={bg=HeatColor!47}, cell{5}{4}={bg=HeatColor!58},
  cell{6}{2}={bg=HeatColor!28}, cell{6}{3}={bg=HeatColor!3}, cell{6}{4}={bg=HeatColor!4},
  cell{7}{2}={bg=HeatColor!100}, cell{7}{3}={bg=HeatColor!100}, cell{7}{4}={bg=HeatColor!100},
}
\toprule
 & de & es & avg. \\ \midrule
\vsix-\cascade   & 16.82 & 13.84 & 15.33 \\
\vsix-\coft      & \textbf{17.76} & \textbf{15.47} & \textbf{16.62} \\
\midrule
\vseven-\cascade & 17.28 & 13.47 & 15.37 \\
\vseven-\coft    & \textbf{18.36} & \textbf{14.99} & \textbf{16.67} \\
\midrule
\veight-\cascade & 17.27 & 13.56 & 15.42 \\
\veight-\coft    & \textbf{18.64} & \textbf{16.66} & \underline{\textbf{17.65}} \\
\bottomrule
\end{tblr}
\label{tab:contraprost}
\end{table}

\section{Results}
\label{sec:results}

\noindent\textbf{\coft models tend to overlook speech inputs.}\;~The interpretability analysis (\S\ref{sec:value_zeroing}) shows that during translation generation, \coft models rely primarily on the transcription and the previously translated tokens. Figure~\ref{fig:input_att} presents a layer-wise view of the contributions. In \vsix, the contribution of speech tokens is close to zero across all layers, whereas \vseven and especially \veight exhibit increased speech attribution in mid-late layers. Averaging the speech contribution across layers confirms these observations: from $0.0228 \pm 0.0005$ in \vsix, the contribution increases to $0.035 \pm 0.001$ (${\sim}1.54\times$) in \vseven, and reaches $0.051 \pm 0.002$ (${\sim}2.24\times$) in \veight. Interestingly, we observe a peak at layer~4 across models, though a detailed analysis of this phenomenon is left to future work. In summary, \coft models internally behave close to \cascade systems, and these training interventions mitigate the over-reliance on the transcript.

\vspace{2pt}

\noindent\textbf{\coft is vulnerable to transcription errors.}\;~The robustness analysis against error propagation (\S\ref{sec:errorprop}) shows that \coft generation behaves similarly to \cascade. Figure~\ref{fig:errprop} illustrates that \vsix performance decreases at almost the same rate for both strategies as noise increases in the transcript. This similarity suggests that \coft, like \cascade, heavily relies on the transcript and largely disregards speech tokens at inference, consistent with our interpretability findings. \vseven is slightly more robust, yielding a shallower slope. The most notable improvement comes with \veight, which displays a much slower performance decline, indicating minimal degradation even when up to 30\% of the words in the transcription are corrupted. While this might suggest that \veight behaves similarly to a \direct system (over-penalizing transcript reliance and losing the advantages of \coft), our interpretability results (Fig.~\ref{fig:input_att}) demonstrate that the model still depends primarily on the transcript. Therefore, this model seems to have an error recovery capability that leverages the speech input to recover from degraded text.

\footnotetext{\hypertarget{linkB}{}Using paired bootstrap resampling (10k resamples) with 95\% CI.}

\stepcounter{footnote}
\newcounter{tablefnA} 
\setcounter{tablefnA}{\value{footnote}}
\footnotetext{\hypertarget{linkA}{}Scores computed with \xcometxl version.}

\begin{table}[ht!]
\centering
\scriptsize
\caption[]{\textit{{\fontsize{5.8pt}{7pt}\selectfont X}{\scriptsize COMET}} (\textuparrow) results on FLEURS (\textit{{\scriptsize EN}\textrightarrow{\scriptsize XX}}) across three model variants and two generation strategies.\protect\hyperlink{linkA}{\footnotemark[\thetablefnA]} Best results are indicated by a more intense {\setlength{\fboxsep}{1pt}\colorbox{HeatColor}{background}}, those with statistically significant\protect\hyperlink{linkB}{\footnotemark[\thetablefnB]} improvements over \textit{B{\scriptsize ASE}} are \underline{underlined}, and \textbf{bold} indicates cases where \textit{C{\scriptsize O}T} significantly outperforms \textit{C{\scriptsize ASCADE}} in the same model variant.}
\resizebox{0.47\textwidth}{!}{
\begin{tblr}{
  width = 0.47\textwidth,
  colspec = {l *{7}{r}},
  column{1} = {leftsep=3pt, rightsep=2pt},
  column{2-8} = {leftsep=3pt, rightsep=3pt},
  row{1} = {halign=c, font=\scshape, abovesep=3pt, belowsep=2pt},
  row{2} = {abovesep=3pt, belowsep=1pt},
  row{3} = {abovesep=1pt, belowsep=2pt},
  row{4} = {abovesep=3pt, belowsep=1pt},
  row{5} = {abovesep=1pt, belowsep=2pt},
  row{6} = {abovesep=3pt, belowsep=1pt},
  row{7} = {abovesep=1pt, belowsep=2pt},
  cell{2}{2}={bg=HeatColor!32}, cell{2}{3}={bg=HeatColor!1}, cell{2}{4}={bg=HeatColor!15}, cell{2}{5}={bg=HeatColor!0}, cell{2}{6}={bg=HeatColor!7}, cell{2}{7}={bg=HeatColor!27}, cell{2}{8}={bg=HeatColor!13},
  cell{3}{2}={bg=HeatColor!0}, cell{3}{3}={bg=HeatColor!0}, cell{3}{4}={bg=HeatColor!0}, cell{3}{5}={bg=HeatColor!16}, cell{3}{6}={bg=HeatColor!0}, cell{3}{7}={bg=HeatColor!0}, cell{3}{8}={bg=HeatColor!0},
  cell{4}{2}={bg=HeatColor!100, cmd=\underline}, cell{4}{3}={bg=HeatColor!100, cmd=\underline}, cell{4}{4}={bg=HeatColor!100, cmd=\underline}, cell{4}{5}={bg=HeatColor!76, cmd=\underline}, cell{4}{6}={bg=HeatColor!76, cmd=\underline}, cell{4}{7}={bg=HeatColor!100, cmd=\underline}, cell{4}{8}={bg=HeatColor!100, cmd=\underline},
  cell{5}{2}={bg=HeatColor!83, cmd=\underline}, cell{5}{3}={bg=HeatColor!84, cmd=\underline}, cell{5}{4}={bg=HeatColor!90, cmd=\underline}, cell{5}{5}={bg=HeatColor!100, cmd=\underline, font=\bfseries}, cell{5}{6}={bg=HeatColor!100, cmd=\underline}, cell{5}{7}={bg=HeatColor!88, cmd=\underline}, cell{5}{8}={bg=HeatColor!100, cmd=\underline},
  cell{6}{2}={bg=HeatColor!35}, cell{6}{3}={bg=HeatColor!11}, cell{6}{4}={bg=HeatColor!41}, cell{6}{5}={bg=HeatColor!9}, cell{6}{6}={bg=HeatColor!38, cmd=\underline}, cell{6}{7}={bg=HeatColor!49}, cell{6}{8}={bg=HeatColor!33, cmd=\underline},
  cell{7}{2}={bg=HeatColor!50, cmd=\underline}, cell{7}{3}={bg=HeatColor!49, cmd=\underline, font=\bfseries}, cell{7}{4}={bg=HeatColor!32, cmd=\underline}, cell{7}{5}={bg=HeatColor!79, cmd=\underline, font=\bfseries}, cell{7}{6}={bg=HeatColor!35, cmd=\underline}, cell{7}{7}={bg=HeatColor!30, cmd=\underline}, cell{7}{8}={bg=HeatColor!49, cmd=\underline, font=\bfseries},
}

\toprule
 & ca & de & es & fr & it & pt & avg. \\ \midrule
\vsix-\cascade   & 86.39 & 92.41 & 87.30 & 81.00 & 87.19 & 87.91 & 87.03 \\
\vsix-\coft      & 85.41 & 92.40 & 86.86 & 81.47 & 87.00 & 87.20 & 86.72 \\
\midrule
\vseven-\cascade & 88.48 & 93.78 & 89.71 & 83.21 & 89.10 & 89.80 & 89.01 \\
\vseven-\coft    & 87.97 & 93.56 & 89.42 & 83.89 & 89.77 & 89.50 & 89.02 \\
\midrule
\veight-\cascade & 86.48 & 92.55 & 88.03 & 81.25 & 88.05 & 88.47 & 87.47 \\
\veight-\coft    & 86.94 & 93.08 & 87.76 & 83.29 & 87.98 & 87.99 & 87.84 \\
\bottomrule
\end{tblr}
}
\label{tab:results}
\end{table}

\vspace{2pt}

\noindent\textbf{Injecting noisy transcripts increases prosody awareness.}\;~As shown in Table~\ref{tab:contraprost}, \coft inference yields some improvement in prosodic sensitivity (\S\ref{subsec:prosody}) across all model variants compared to \cascade, which is expected since \cascade does not have access to acoustic cues from the source speech, resulting in relatively consistent scores across variants. Importantly, \veight-\coft achieves the highest scores and substantially widens the gap between \coft and \cascade compared to \vsix and \vseven, whose benefits are limited, revealing that there is still margin for improvement in the prosody awareness of CoT S2TT systems.

\vspace{2pt}

\noindent\textbf{Our interventions do not harm translation quality.}\;~We study methods to increase speech input awareness, which transfer to better robustness to error propagation and improved use of prosodic information. However, these interventions would not be helpful if they harmed performance in general translation benchmarks. As shown in Table~\ref{tab:results}, both interventions yield gains over \vsix, particularly \vseven. This aligns with the common practice in the field of mixing Direct and CoT samples in training data~\cite{huang2023speech}, providing new insights into why this mixture might be crucial. Our interventions also allow \coft to close the gap with \cascade, achieving comparable performance in \vseven and surpassing it in \veight.

\vspace{9pt}

\section{Conclusions}
\label{sec:Conclusion}

We analyzed CoT S2TT beyond general translation quality by examining input attributions, robustness to transcription errors, and prosody awareness. While it is often assumed that the joint availability of speech tokens and transcripts allows the model to leverage the speech signal directly, our results show that CoT heavily resembles a cascade system in practice. It relies primarily on transcripts, remains vulnerable to error propagation, and yields only limited improvements in prosody awareness compared to the cascaded approach.

Our experiments demonstrate that while introducing Direct S2TT data into the training mixture (\vseven) improves overall translation quality, speech utilization remains limited. Conversely, injecting noisy transcripts during training (\veight) yields marginal improvements on general benchmarks, but significantly enhances speech awareness. This results in much better robustness to error propagation and increased sensitivity to prosody. Future work could explore whether combining \vseven and \veight strategies yields complementary benefits.

This work provides an initial study of CoT S2TT behavior in scenarios not fully captured by general translation benchmarks. Our analysis of different training interventions offers practical insights into how these systems can be improved, moving toward S2TT models that can integrate more effectively speech and text signals.

\vfill\pagebreak

\section{Generative AI Use Disclosure}

The code used along this article was developed with the assistance of GitHub Copilot and Cursor as programming aids. Training data for the \veight configuration and the corrupted transcripts used to assess the robustness to error propagation were generated with Gemini 2.0 Flash. Lastly, ChatGPT and Gemini were used to improve spelling, grammar, clarity, and overall readability during manuscript preparation. 

\section{Acknowledgments}
\label{sec:Acknowledgments}

This work is funded by the Ministerio para la Transformación Digital y de la Función Pública and Plan de Recuperación, Transformación y Resiliencia - Funded by EU – NextGenerationEU within the project Desarrollo Modelos ALIA. Project AIA2025-163322-C66 funded by MICIU/AEI/10.13039/501100011033. Federico Costa and Cristina España-Bonet acknowledge their AI4S fellowship within the ``Generación D'' initiative by Red.es, Ministerio para la Transformación Digital y de la Función Pública, for talent attraction (C005/24-ED CV1), funded by NextGenerationEU through PRTR.

\bibliographystyle{IEEEtran}
\bibliography{mybib}

\end{document}